\documentclass[10pt,twocolumn,letterpaper]{article}
\usepackage{cvpr}              

\usepackage[breaklinks,colorlinks,allcolors=cvprblue]{hyperref}
\usepackage{algorithm}
\usepackage{algorithmic}
\usepackage{stfloats}
\usepackage[accsupp]{axessibility}  
\usepackage{booktabs}
\usepackage{array}
\usepackage{float}

\definecolor{cvprblue}{rgb}{0.21,0.49,0.74}

\def\paperID{AI4RWC-54} 
\def\confName{CVPR}
\def\confYear{2026}

\title{MOTOR-Bench: A Real-world Dataset and  Multi-agent Framework for Zero-shot Human Mental State Understanding}

\author{Xiaoyu Yuan\\
University of Oulu\\
{\tt\small Xiaoyu.Yuan@oulu.fi}
\and
Niklas Heikkala\\
University of Oulu\\
{\tt\small niklas.heikkala@oulu.fi}
\and
Tiina Törmänen\\
University of Oulu\\
{\tt\small tiina.tormanen@oulu.fi}
\and
Hanna Järvenoja\\
University of Oulu\\
{\tt\small hanna.jarvenoja@oulu.fi}
\and
Guoying Zhao\\
University of Oulu\\
{\tt\small guoying.zhao@oulu.fi}
\and
Haoyu Chen\\
University of Oulu\\
{\tt\small chen.haoyu@oulu.fi}
}

\begin{document}
\maketitle
\thispagestyle{empty}
\begin{abstract}
Understanding human mental states from natural behavior is crucial for intelligent systems in the real world. However, most current research focuses on predicting isolated mental state labels, lacking structured annotations of complex interpersonal interactions.
To support structured analysis, we introduce \textbf{MOTOR-Bench}, a carefully-designed benchmark with a real-world dataset \textbf{MOTOR-dataset}, containing 1,440 multimodal video clips in collaborative learning scenarios, reflecting key real-world data challenges including natural class imbalance, visual noise, and domain-specific language. Each sample is labeled by educational experts based on self-regulated learning theory. We further evaluate several state-of-the-art multimodal large language models and multi-agent systems in a zero-shot setting on our MOTOR-Bench. However, their performance on this task remains limited, suggesting that existing methods still struggle with structured reasoning from observable behavior to deeper mental states. To address this challenge, we propose a reasoning multi-agent framework, named \textbf{MOTOR-MAS}. It coordinates multiple agents through a structured agent coordination mechanism to infer explicit behaviors, internal cognitions, and psychological emotions.
Experimental results show that our MOTOR-MAS outperforms the best single-model benchmark by 15.93 points in Macro-F1 scores for the three labels of behavior, cognition, and emotion, and outperforms the general multi-agent benchmark by 10.2 points in internal cognition prediction.
\end{abstract}






\section{Introduction}

\begin{figure}[t]
  \centering
  \includegraphics[width=0.98\linewidth]{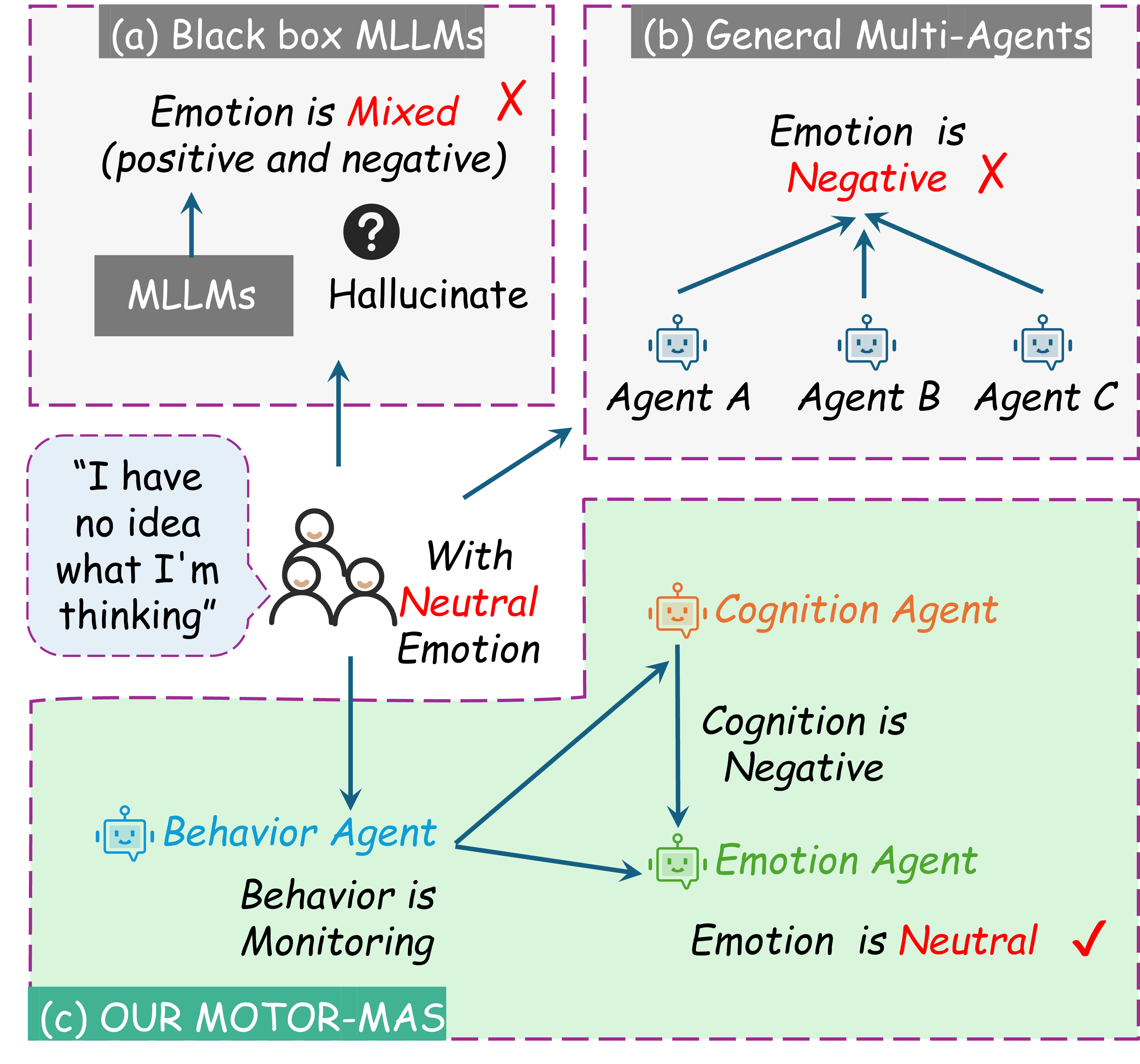} %
  \caption{Three paradigms for inferring visual mental states.
  (a) Black-box multimodal language models typically rely on surface-level visual cues.
  (b) General multi-agent frameworks assign tasks to specific agents but lack structured coordination among agents.
  (c) MOTOR-MAS integrates reasoning across behavior, cognition, and emotion for more reliable mental state inferences.}
  \label{fig:cover}
\end{figure}

Inferring human mental states from natural behavior is important, but difficult, for intelligent systems in real-world settings~\cite{yang2026saynext,lian2025emoprefer,lian2024ov,zhao_li_xu_2022}. In education, social interaction, and healthcare, people communicate through speech, gestures, and facial expressions while also expressing thoughts and emotions~\cite{zafeiriou2016facial,guerdelli2025multimodal,chen2023smg,liu2021imigue,chen2019analyze}. Yet \textit{outward behavior does not always reveal what a person is thinking or feeling}, which is a long-standing problem in the field~\cite{mittal2020emoticon,srivastava2023you}. For example, someone may smile while saying, \textit{``I really have no idea what I'm thinking."} In this case, the visible cue is the smile, which may appear positive at first glance. However, the utterance itself expresses confusion and uncertainty, suggesting a negative cognitive state. These signals do not align: a seemingly positive expression can co-occur with a negative underlying cognitive state. This makes mental state understanding more challenging than predicting a single isolated label.

Thus, inspired by the success of learning science theories in explaining collaborative learning processes \cite{jarvenoja2026motivational,jarvenoja2025succeed,sobocinski2020}, we study mental state inference in collaborative learning environments by introducing a new benchmark, named \textbf{MOTOR-Bench}. Specifically, due to the inherent nature of mental states (abstract and difficult to define) in general settings, we choose to study them in a concrete interaction context where participants naturally express understanding, confusion, and intentions through multimodal behaviors. Based on this idea, we introduce the \textbf{MOTOR-dataset}, a real-world dataset consisting of 1,440 multimodal video clips collected from classroom collaborative learning activities. Unlike most existing work that predicts only one type of mental state label from visual or multimodal signals, we annotate each sample with three complementary dimensions: \textit{behavior}, \textit{cognition}, and \textit{emotion}. The dataset also reflects practical challenges such as visual noise, class imbalance, and domain-specific language.

We evaluate several state-of-the-art multimodal large language models (MLLMs), including Gemini ~\cite{gemini}, InternVL ~\cite{internvl,wang2025internvl3}, and AffectGPT~\cite{affectgpt}, as well as the general multi-agent framework CAMEL~\cite{li2023camel} on our MOTOR-bench, in a zero-shot setting. Although these models perform well on many vision-language tasks, they struggle with real-world mental state inference. On the MOTOR-Bench, the best-performing MLLM, InternVL-3.5~\cite{wang2025internvl3}, reaches only 26.84 Macro-F1~\cite{opitz2019macro} averaged over behavior, cognition, and emotion prediction. CAMEL performs better overall, with 39.28 Macro-F1, but still achieves only 29.61 on cognition prediction. These results show that specific mental state inference remains difficult in real-world interactions as shown in \cref{fig:cover}.

Thus, we further propose \textbf{MOTOR-MAS}, a customized reasoning-based multi-agent framework for structured mental state inference. Instead of predicting each dimension independently, MOTOR-MAS predicts behavior, cognition, and emotion through coordinated specialized agents. Intermediate predictions from earlier stages are used to support subsequent predictions, enabling structured reasoning across behavior, cognition, and emotion.

Experiments show that our MOTOR-MAS improves performance substantially on the MOTOR benchmark. It achieves a Macro-F1 score of 42.77, outperforming the strongest single-model baseline by 15.93 points. Compared with the general multi-agent baseline, it also improves cognition prediction by 10.2 points. These results suggest that structured coordination among specialized agents is effective for mental state inference in real-world collaborative learning scenarios.
\section{Related Work}
\label{sec:related}

\subsection{Mental State Understanding in AI}

Mental state understanding in AI is commonly studied by inferring latent human states from observable signals such as language, facial expression, speech, and interaction context~\cite{picard2000affective, poria2017review}. Early work mainly focused on affective computing and unimodal or weakly multimodal cues, showing that internal states can often be approximated from external behavior, but usually along a single dimension such as emotion or sentiment~\cite{picard2000affective, poria2017review}.

Recent work has shifted toward richer multimodal settings that combine text, audio, and visual information. Datasets such as CMU-MOSEI support multimodal sentiment and emotion analysis in open-domain videos~\cite{zadeh2018multimodal}. MELD further models emotion in multi-party conversation and highlights the importance of contextual interaction cues~\cite{poria2019meld}. These studies are closely related to our task, but most still focus on predicting one state dimension at a time. In contrast, our setting models behavior, cognition, and emotion together as a structured inference problem in collaborative learning.

\subsection{Multi-Agent Frameworks: Structure or Scale}

In multi-agent architectures, role specialization is more effective than scaling. Nevertheless, how to classify the roles of the agents and how the agents interact still require further exploration. CAMEL ~\cite{li2023camel} shows that assigning different dialogue roles can achieve collaborative problem-solving beyond single-agent benchmarks, while MetaGPT ~\cite{hong2023metagpt} further structures agent communication through standardized operational procedures. These frameworks demonstrate that collaboration is helpful, but their structure stems from heuristics of task decomposition rather than domain theory. Increasing the number of agents does not solve this limitation. Qian et al. discovered that performance follows a reasonable growth pattern and tends to saturate before reaching its theoretical limit~\cite{qian2024scaling}. Yang et al. ~\cite{yang2026understanding} provides a formal explanation: homogeneous agents produce correlated outputs, thus the evidence contributed by newly added agents diminishes. Two different agents can achieve the same processing power as sixteen homogeneous agents, indicating that the structure of information is far more important than the number of agents. 

However, these frameworks are not designed specifically for structured mental state understanding. In particular, they do not provide task-specific mechanisms for coordinating information across behavior, cognition, and emotion. MOTOR-MAS addresses this problem by allowing domain knowledge to guide the design of agent communication: concepts related to self-regulated learning guide how information flows between agents.

\subsection{Learning Science for Mental State Analysis}

Learning science research has examined how behavior, cognition, and emotion interact in collaborative learning. Work on self-regulated learning, in particular, describes how learners monitor progress, evaluate their situation, and adjust their actions during task-oriented interaction~\cite{zimmerman2002becoming}. Educational researchers have also developed coding schemes that label collaborative episodes in terms of behavioral, cognitive, and emotional dimensions~\cite{azevedo2010does, jarvenoja2025succeed, jarvenoja2026motivational}. These studies are closely related to mental state analysis, but they are rarely translated into AI benchmarks or multimodal reasoning systems for real-world scenarios. This has led to a gap between learning-based annotation frameworks and computational systems for natural collaborative interactions. 

Our work helps bridge this gap by introducing the MOTOR-Bench, a real-world benchmark for behavioral, cognitive, and emotional analysis, and the MOTOR-MAS, a structured multi-agent framework tailored for this scenario.

\section{The MOTOR-Bench}

\begin{figure*}[t]
    \centering
      \includegraphics[width=0.98\linewidth]{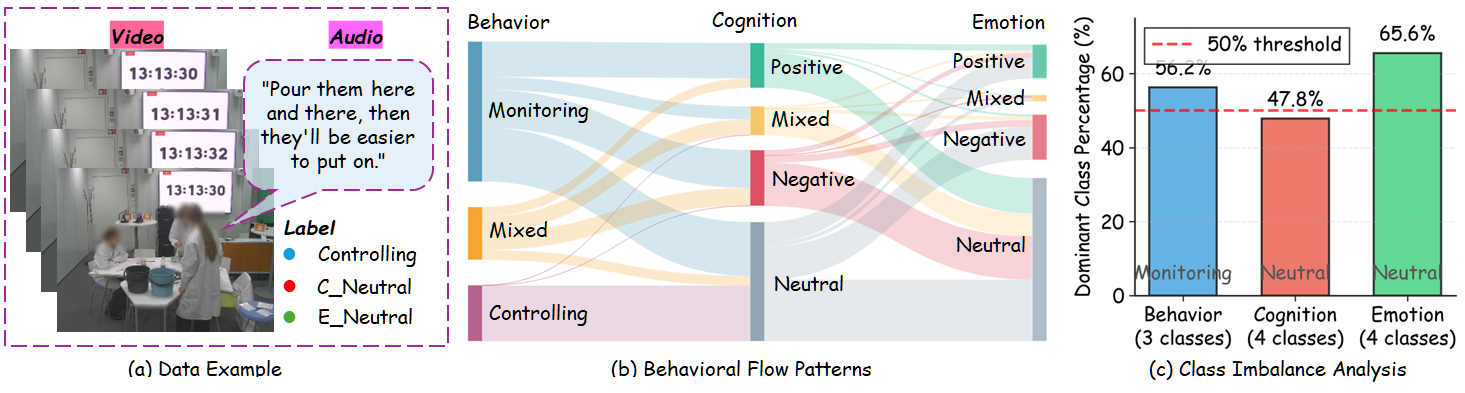} 
    \caption{Dataset characteristics and flow analysis. (a) Example of a collaborative learning segment with video frames, transcript, and behavior--cognition--emotion labels. (b) Flow patterns across behavior, cognition, and emotion are shown with Sankey diagrams. (c) Class distributions across the three dimensions. }
\label{fig:dataset}
\end{figure*}

Most multimodal datasets for emotion or mental state analysis focus on general-domain settings such as movies or clinical interviews. These settings differ substantially from collaborative learning. In collaborative learning environments, people interact more frequently, and their emotions are more subtle. To study mental state inference in this setting, we build the \textbf{MOTOR-Bench}~\cite{jarvonoja2024motor,jarvenoja2025succeed,jarvenoja2026motivational}, a real-world benchmark collected from classroom collaboration.

\subsection{Task Definition and Label Space}

We formulate mental state understanding in collaborative learning as a structured prediction task over three dimensions: behavior, cognition, and emotion. Following prior SRL-based coding frameworks~\cite{zimmerman2002becoming, jarvenoja2026motivational}, we define a label space that captures both observable interactions and internal mental states.

The \textbf{behavior dimension} $B$ is derived from the behavioral coding axis in the original framework. Here, \textit{Monitoring} refers to interactions that increase awareness of the group’s current state, including task understanding, progress, and ongoing emotional and motivational conditions. \textit{Controlling} refers to attempts to influence or regulate these states. Episodes that contain inseparable elements of both are labeled as \textit{Mixed}. We therefore define $B \in \{\text{Monitoring}, \text{Controlling}, \text{Mixed}\}$.

The \textbf{cognition} and \textbf{emotion} dimensions describe internal aspects of regulation ~\cite{zimmerman2002becoming}. The \textbf{cognition dimension} $C$ captures the learner’s metacognitive evaluation of the ongoing task. It indicates whether the learner expresses a positive, negative, mixed, or absent evaluation. We define $C \in \{\text{Positive}, \text{Negative}, \text{Mixed}, \text{Neutral}\}$.
The \textbf{emotion dimension} $E$ captures the affective state expressed during collaboration. It reflects whether the learner’s emotion is positive, negative, mixed, or neutral. We define $E \in \{\text{Positive}, \text{Negative}, \text{Mixed}, \text{Neutral}\}$.

These three dimensions preserve the structure of the original annotation framework while making the output space suitable for structured prediction.

\subsection{Data Collection}

\Cref{fig:dataset}(a) shows an example from the MOTOR-dataset. Each sample contains synchronized video frames, a Finnish transcript, and behavior--cognition--emotion labels. The dataset includes 1,440 video clips collected from a nature-based collaborative learning activity at a university in a European country~\cite{jarvonoja2024motor}. The participants were 30 groups of secondary school students engaged in inquiry-based science experiments. Each clip contains about 6.10 seconds of interaction between 2 to 4 students.

All communication was in Finnish. We transcribed the recordings using Whisper-large-finnish-v3~\cite{finnish-nlp_2023}. On standard Finnish speech benchmarks, CSS10-fi~\cite{park2019css10} and VoxPopuli-fi~\cite{wang2021voxpopuli}, this model achieves a word error rate of 14.26\%, which we found sufficient for further analysis. All participants, legal guardians, and relevant municipal departments provided informed consent. The study received approval from the relevant institutional ethics review board. Video data was anonymized before analysis. We detected faces with OpenFace 2.0 and applied Gaussian blurring so that no personally identifiable information was retained.

\subsection{Annotation Scheme}

Each clip is annotated with a behavior--cognition--emotion triplet following prior SRL-based coding schemes~\cite{jarvenoja2025succeed,jarvenoja2026motivational}. Trained educational researchers first labeled a subset of the data independently, then compared results, resolved disagreements through discussion, and refined the category descriptions before annotating the full dataset.

Behavior labels indicate whether an episode reflects \textit{Monitoring}, \textit{Controlling}, or \textit{Mixed}. Cognition labels describe the valence of metacognitive evaluation: \textit{C\_Positive}, \textit{C\_Negative}, \textit{C\_Mixed}, or \textit{C\_Neutral}~\cite{jarvenoja2026motivational}. Emotion labels are \textit{E\_Positive}, \textit{E\_Negative}, \textit{E\_Neutral}, and \textit{E\_Mixed}. Inter-rater reliability was measured on 25\% of the data using Cohen's kappa. We obtained good agreement for monitoring and controlling behavior ($\kappa = 0.73$) and for the valence of monitoring ($\kappa = 0.70$).

\subsection{Dataset Statistics and Real-world Challenges}


The MOTOR dataset is highly imbalanced, reflecting real collaborative learning. As shown in \cref{fig:dataset}(c), the dominant classes in behavior, cognition, and emotion are monitoring (56.2\%), neutral (47.8\%), and neutral (65.6\%), respectively.

This pattern is not an artifact of collection. In classroom collaboration, most interaction is task-oriented and emotionally neutral. Strong evaluative or affective states appear only in part of the discussion. This makes the MOTOR-dataset quite different from entertainment-oriented datasets, where emotionally salient events are much more frequent.

\Cref{fig:dataset}(b) also shows clear cross-dimensional patterns. Controlling behavior is strongly associated with \textit{C\_Neutral} (99.1\% of controlling instances), while monitoring behavior is distributed across several cognitive states: \textit{C\_Neutral} (38.4\%), \textit{C\_Negative} (27.0\%), \textit{C\_Positive} (25.2\%), and \textit{C\_Mixed} (9.4\%). We also observe a clear relation between cognition and emotion: \textit{C\_Negative} is often paired with \textit{E\_Negative}, whereas \textit{C\_Neutral} most often co-occurs with \textit{E\_Neutral}. These patterns motivate structured reasoning across the three dimensions.

\section{Methodology}
\label{sec:method}

We formulate visual mental state understanding as a structured reasoning problem based on self-regulated learning theory~\cite{zimmerman2002becoming}, and then implement this framework through a multi-agent architecture.

\subsection{Motivation for Structured Reasoning}

\label{sec:srl_theory} 
Previous research on collaborative learning has shown that there is a close connection between behavior, cognition, and emotion, rather than fully independent ~\cite{zimmerman2002becoming}. Observable behaviors such as monitoring and control often provide useful contextual information for interpreting underlying cognitive states, while emotional responses are often associated with behavioral context and cognitive cues.

Based on this perspective, we organize the inference of mental states into a structured reasoning process that spans three dimensions. 
We first infer behavior based on multimodal observations, then use behavioral cues to support cognitive predictions, and finally predict emotions based on both behavior and cognition inputs.
This structural sequence design provides a simple and interpretable approach to modeling the dependencies between behavior, cognition, and emotion in collaborative learning scenarios.

\subsection{MOTOR-MAS: Anchor-and-Derive}
\label{sec:model}

Given a collaborative learning video clip $V$ and its transcript $T$, our MOTOR-MAS predicts a structured output tuple $Y=(B,C,E)$, where $B$, $C$, and $E$ denote behavior, cognition, and emotion, respectively. As introduced in Section~\ref{sec:srl_theory}, these three dimensions are closely related in collaborative learning and are not equally observable from raw multimodal input. In particular, behavior is relatively explicit, whereas cognition and emotion are often more subtle and context-dependent. This motivates a staged inference process, in which intermediate predictions from earlier stages are used to support later ones.

\begin{figure}[t]
  \centering
  \includegraphics[width=0.98\linewidth]{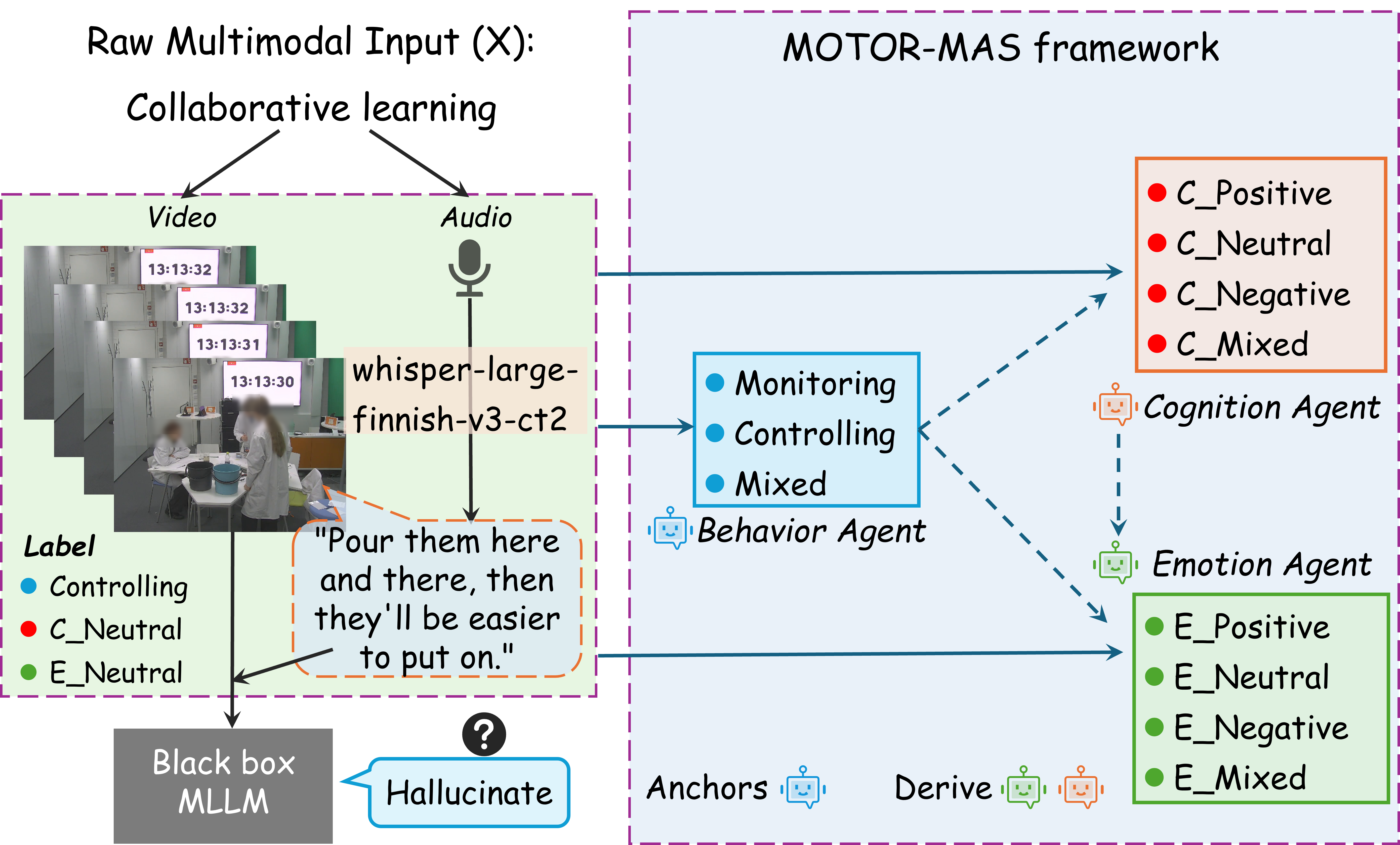} %
    \caption{Overview of the MOTOR-MAS. Multimodal input (video and text) is processed sequentially by three specialized agents. The Behavior Agent first predicts observable actions, the Cognition Agent infers cognitive states based on behavior and multimodal context, and the Emotion Agent predicts emotion using both behavior and cognition. The dashed box contrasts this structured reasoning process with a black-box MLLM baseline that predicts directly from raw input.}
  \label{fig:framework} 
\end{figure}

We implement this staged reasoning process as a structured sequential decomposition over behavior, cognition, and emotion, as shown in \cref{fig:framework}. Unlike generic sequential prompting, each agent is equipped with SRL-grounded task specifications that encode the theoretical role of its assigned dimension, making the information passed between agents semantically structured rather than arbitrary intermediate outputs.
\begin{equation}
X \rightarrow B, \quad \{X, B\} \rightarrow C, \quad 
\{X, B, C\} \rightarrow E,
\label{eq:scm}
\end{equation}
where $X = \{V, T\}$ denotes multimodal observations. This formulation reflects a staged reasoning process across the three dimensions. Behavior is inferred directly from multimodal input. Cognition is predicted using both the input and the inferred behavior. Emotion is then predicted using the combined context of input, behavior, and cognition. 
Decomposing the joint distribution $P(B,C,E|X)$ according to this ordering yields
\begin{equation}
P(C|X) = \sum_{B} P(C|B,X)\, P(B|X),
\end{equation}
\begin{equation}
P(E|X) = \sum_{B,C} P(E|C,B,X)\, P(C|B,X)\, P(B|X).
\end{equation}
This decomposition method directly corresponds to three specialized agents, each responsible for a conditional probability in the structured sequence, as shown in \cref{fig:framework}.

The \textbf{Behavior Agent} serves as the anchor of the entire framework by estimating $P(B|X)$ to isolate overt learning actions from multimodal input. It categorizes actions into monitoring, controlling, or a combination of both by prioritizing directional intent over surface-level linguistic features. This mechanism functions as an information bottleneck that filters transient visual noise and prevents spurious associations between raw perceptual signals and latent mental states.

The \textbf{Cognition Agent} infers latent evaluative states 
that are not directly observable from raw input by computing 
$P(C|B,X)$ conditioned on the behavioral anchor provided by the Behavior Agent. It characterizes the valence of metacognitive judgment as positive, negative, mixed, or neutral. Grounding this inference in $B$ reduces the ambiguity inherent in interpreting covert cognitive processes from noisy multimodal observations.

The \textbf{Emotion Agent} performs structured reasoning by calculating $P(E|C,B,X)$, which comprehensively considers behavioral anchoring factors and results based on cognitive assessment, thereby classifying emotional states.
Given the task-oriented nature of collaborative learning, the agent adopts a neutrality-default assumption that is overridden only when explicit emotional markers are present in the 
transcript or visual features. As shown in \cref{fig:framework}, this disentangled design ensures that each prediction is consistent with earlier intermediate predictions rather than 
directly hallucinated from raw visual input.

\section{Experiments}
\label{sec:experiments}

We evaluate MOTOR-MAS through benchmark comparison and ablation studies.

\subsection{Experimental Setup}

We compare MOTOR-MAS with several strong baselines for multimodal mental state inference. These include InternVL-2.5~\cite{internvl} and InternVL-3.5~\cite{wang2025internvl3} as representative open-source MLLMs, Gemini-1.5-Flash and Gemini-1.5-Pro~\cite{gemini} as proprietary frontier models, and AffectGPT~\cite{affectgpt} as a domain-specific model for affective computing. We also conducted comparative experiments using the general multi-agent framework CAMEL ~\cite{li2023camel}. All baselines are evaluated in a zero-shot setting with standardized prompts.

For the CAMEL baselines, we use the InternVL-3.5-8B checkpoint \footnote{\url{https://huggingface.co/OpenGVLab/InternVL3_5-8B}} as the shared backbone to keep the comparison aligned with MOTOR-MAS. In the 2-agent setting, two agents jointly discuss and predict the labels. In the 3-agent setting, three agents are assigned to behavior, cognition, and emotion prediction, respectively. All CAMEL variants receive the same multimodal input and use the same decoding settings as the MOTOR-MAS.

Our MOTOR-MAS also uses the InternVL-3.5-8B checkpoint as the backbone for all three agents. The model runs in bfloat16 with Flash Attention~\cite{dao2023flashattention2}. For each sample, we uniformly select eight video frames and resize them to $448 \times 448$ following the InternVL preprocessing pipeline~\cite{wang2025internvl3}. 
We integrate Finnish transcripts generated by Whisper-large-finnish-v3~\cite{finnish-nlp_2023} into task-specific prompts incorporating SRL-inspired instructions.

We use greedy decoding with a temperature of 0 for all agents. The maximum generation length is 1024 tokens. Final labels for behavior $B$, cognition $C$, and emotion $E$ are extracted from the generated text using regular expressions over the predefined label set.

We report Macro-F1 as the primary metric because the dataset is highly imbalanced. Accuracy is included as a secondary metric. For each task, we compute per-class precision and recall and then macro-average across classes. All experiments are run on NVIDIA V100 GPUs.

\subsection{Main Results}

\Cref{tab:main_results} reports the main results on the MOTOR-Bench. Our MOTOR-MAS achieves 42.77 Macro-F1, outperforming the strongest single-model baseline, InternVL-3.5, by 15.93 points. This result shows that black-box MLLMs still struggle with structured mental state inference in real-world interactions. General multi-agent collaboration helps, but it is still not enough. CAMEL with three agents reaches 39.28 Macro-F1, but remains much weaker on cognition prediction than MOTOR-MAS (29.61 vs.\ 39.81). This suggests that simply adding more agents is not sufficient. How intermediate information is organized also matters.

Performance also differs across the three dimensions. Behavior is the easiest to predict, with 45.23 Macro-F1, likely because many behavioral cues are explicit in the transcript. Emotion follows at 43.27. Cognition remains the hardest dimension at 39.81, which is consistent with its more latent and context-dependent nature. Overall, these results suggest that structured reasoning is especially useful when the target state is not directly observable.

\begin{table}[h]
\centering
\caption{Comparison with state-of-the-art methods on the MOTOR dataset (Macro-F1 score). Our MOTOR-MAS is based on InternVL-3.5. The best performances are marked in bold.}
\label{tab:main_results}
\resizebox{\columnwidth}{!}{%
\begin{tabular}{@{}lccc|c@{}}
\toprule
Method & Behavior & Cognition & Emotion & Avg \\
\midrule

\multicolumn{5}{@{}c}{\textbf{MLLMs}} \\
InternVL-2.5~\cite{internvl} & 24.71 & 16.60 & 31.19 & 24.17 \\
InternVL-3.5~\cite{wang2025internvl3} & 25.89 & 19.21 & 35.43 & 26.84 \\
AffectGPT~\cite{affectgpt} & 22.84 & 11.64 & 33.32 & 22.60 \\
Gemini-1.5-Flash~\cite{gemini} & 24.59 & 20.27 & 23.21 & 22.69 \\
Gemini-1.5-Pro~\cite{gemini} & 25.57 & 21.71 & 22.22 & 23.17 \\

\midrule
\multicolumn{5}{@{}c}{\textbf{Multi-Agent Frameworks}} \\
CAMEL~\cite{li2023camel} (2 Agents) & 41.99 & 24.83 & 28.70 & 31.84 \\
CAMEL~\cite{li2023camel} (3 Agents) & 42.62 & 29.61 & 45.60 & 39.28 \\
MOTOR-MAS (Ours) & \textbf{45.23} & \textbf{39.81} & \textbf{43.27} & \textbf{42.77} \\

\bottomrule
\end{tabular}
}
\end{table}

\subsection{Ablation Study}

\begin{figure}
  \centering
  \includegraphics[width=0.99\columnwidth]{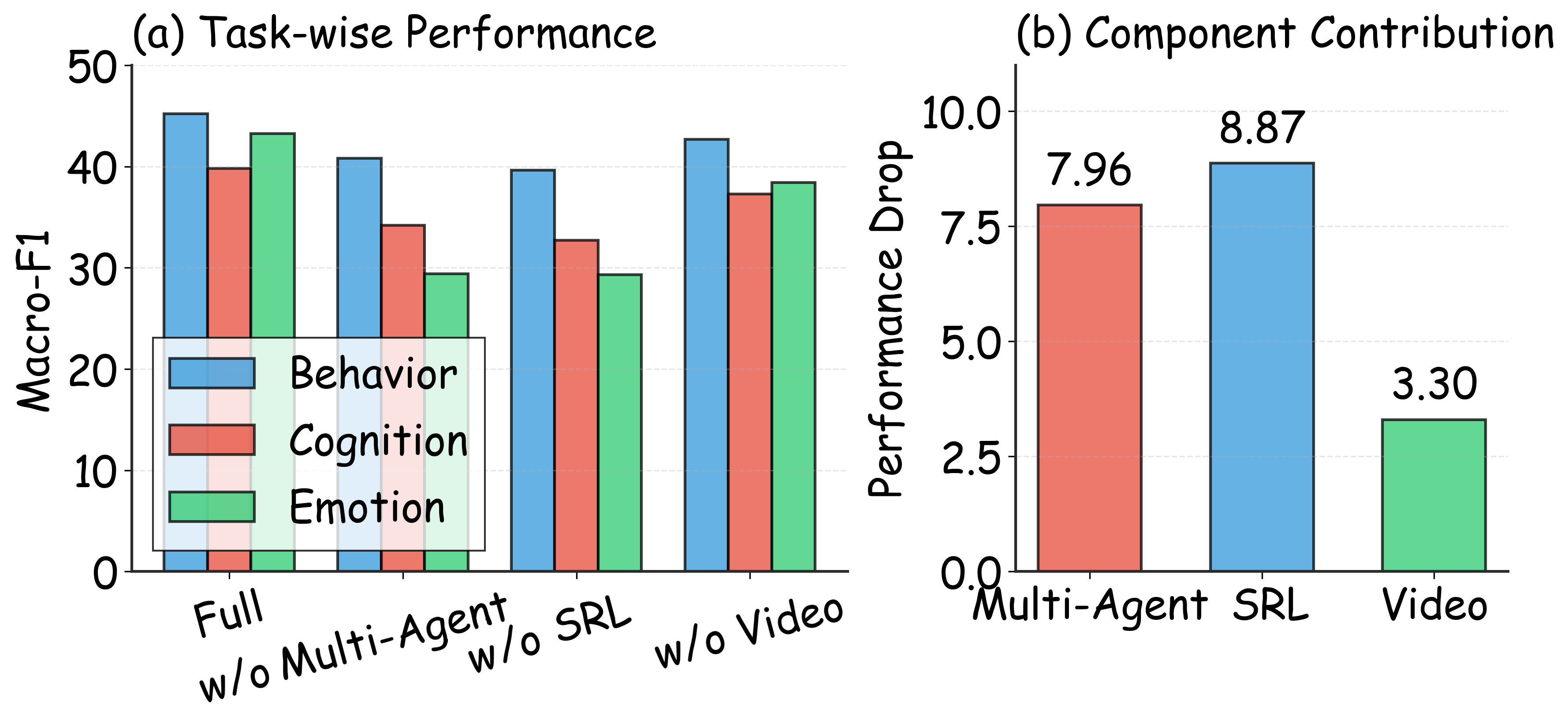}
  \caption{Ablation results. (a) Per-task Macro-F1 across 
model variants. (b) Performance drop from removing each 
component; multi-agent structure and SRL priors contribute 
comparably (${\sim}8$--$9\%$ each), while video input 
accounts for a modest $3.3\%$.}
  \label{fig:ablation}
\end{figure}

We ablate the main components of the MOTOR-MAS to understand where the performance gain comes from. The results are shown in \cref{tab:ablation} and \cref{fig:ablation}.

\textbf{w/o Multi-Agent.}
To test the role of the multi-agent design, we remove the agent-level decomposition and ask a single model to predict behavior, cognition, and emotion within one reasoning process. This variant still keeps the sequential order $B \rightarrow C \rightarrow E$, but no longer separates the task into specialized agents. Performance drops from 42.77 to 34.81 Macro-F1, a decrease of 7.96 points. The largest drops appear in cognition (5.59 points) and emotion (13.88 points), suggesting that agent specialization is especially helpful for latent state inference. Behavior also drops by 4.40 points, which indicates that even relatively explicit tasks benefit from dedicated reasoning rather than joint prediction in a single pass.

\textbf{w/o SRL Priors.}
To examine the effect of domain knowledge, we remove the SRL-inspired instructions from the prompts and replace them with generic task descriptions. This reduces the average Macro-F1 from 42.77 to 33.90, a drop of 8.87 points. The degradation is most visible in behavior (5.58 points) and cognition (7.08 points). This result suggests that prompt-level domain guidance is important for this task, especially when the model needs to infer cognitive states from subtle evidence. The similar performance drops of w/o Multi-Agent and w/o SRL Priors also indicate that structure and domain guidance contribute in complementary ways.

\textbf{w/o Video.}
We also test a text-only variant to measure the contribution of visual input. Removing video reduces the average Macro-F1 from 42.77 to 39.47, a drop of 3.30 points. The effect is smallest for cognition (2.51 points), moderate for behavior (2.55 points), and largest for emotion (4.83 points). This suggests that transcripts carry much of the useful signal, while visual cues remain particularly helpful for emotion prediction. The relatively small drop is also consistent with the nature of the dataset, where classroom videos are often noisy and low-resolution, making text the stronger modality in many cases.

\begin{table}
\centering
\caption{Ablation study of the MOTOR-MAS components. $\Delta$ represents the drop in Average Macro-F1 compared to the Full Model.}
\label{tab:ablation}

\resizebox{\columnwidth}{!}{
\begin{tabular}{@{}lccccc@{}}
\toprule
Variant & Behavior & Cognition & Emotion & Avg & $\Delta$ \\
\midrule
Full Model & 45.23 & 39.81 & 43.27 & 42.77 & - \\
w/o Multi-Agent & 40.83 & 34.22 & 29.39 & 34.81 & $-7.96$ \\
w/o SRL & 39.65 & 32.73 & 29.32 & 33.90 & $-8.87$ \\
w/o Video & 42.68 & 37.30 & 38.44 & 39.47 & $-3.30$ \\
\bottomrule
\end{tabular}
}
\end{table}

\begin{figure*}[ht]
  \centering
    \includegraphics[width=0.98\linewidth]{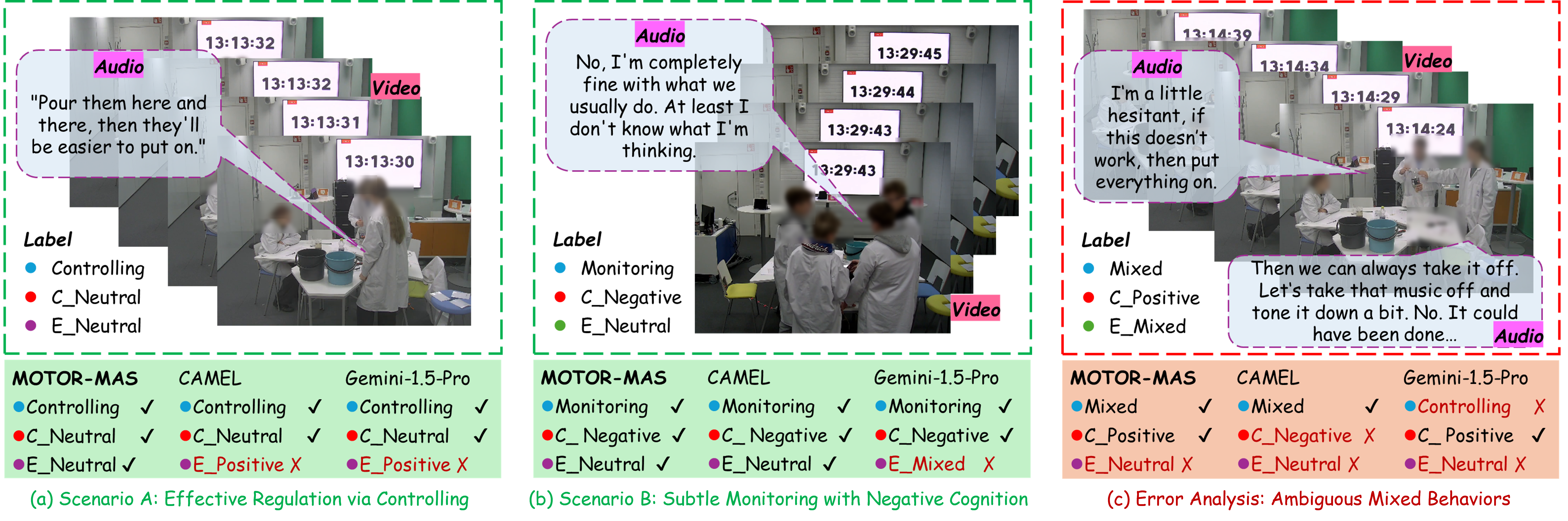} 

  \caption{Qualitative comparison on three cases. 
\textcolor{green}{Green}/\textcolor{red}{red} indicate correct/incorrect MOTOR-MAS predictions. 
(a) Standard controlling behavior: baselines conflate cooperation with positive emotion. 
(b) Monitoring with negative cognition: MOTOR-MAS correctly predicts Neutral emotion; Gemini-1.5-Pro hallucinates Mixed. 
(c) Ambiguous mixed behavior: all models fail on fine-grained mixed emotion.}
\label{fig:case_study}
\end{figure*}

\subsection{Qualitative Analysis}

\Cref{fig:case_study} shows three representative cases that help explain where the MOTOR-MAS succeeds and where it still fails.

\textbf{Case (a): Standard Control Behavior.}
The student says, ``\textit{Pour them here and there, then they’ll be easier to put on.}" All three models correctly identify the behavior as control. However, CAMEL and Gemini-1.5-Pro both predict \textit{E\_Positive}, apparently treating task-oriented cooperation as positive emotion. Our MOTOR-MAS instead predicts \textit{E\_Neutral}, which better matches the context. This case shows that even when behavior is easy to identify, emotion prediction can still fail if the model relies too directly on surface cues.

\textbf{Case (b): Monitoring with Negative Cognition.}
The student smiles while saying, ``\textit{No, I’m completely fine with what we usually do. At least I don’t know what I’m thinking.}" The MOTOR-MAS correctly predicts \textit{E\_Neutral}. Here the model does not simply map negative cognitive content to negative emotion. Instead, it uses the behavioral and cognitive context together. CAMEL reaches the same prediction, while Gemini-1.5-Pro predicts \textit{E\_Mixed}, likely reacting to the smile and the negative statement at the same time without separating their roles in the interaction.

\textbf{Case (c): Ambiguous Mixed Behavior.}
The student first hesitates and then proposes a fallback plan: ``\textit{I’m a little hesitant, if this doesn’t work, then put everything on. Then we can always take it off…}" MOTOR-MAS correctly identifies the behavior as mixed and predicts \textit{C\_Positive}, capturing the recovery-oriented part of the utterance. CAMEL also detects mixed behavior, but predicts \textit{C\_Negative}, focusing too much on the initial hesitation. Gemini-1.5-Pro predicts pure control and misses the evaluative component altogether. All three models fail on \textit{E\_Mixed}, which suggests that fine-grained mixed emotions remain difficult even when behavior and cognition are handled reasonably well.

Overall, these cases suggest that MOTOR-MAS is strongest when later predictions can benefit from earlier intermediate cues. It is especially helpful for reducing obvious mismatches between behavior, cognition, and emotion. At the same time, subtle mixed emotional states remain challenging and point to an important direction for future work.

\section{Conclusion and Limitations}

In this paper, we introduced \textbf{MOTOR-Bench}, a real-world multimodal benchmark for structured mental state inference in collaborative learning. 
The dataset provides annotations for behavior, cognition, and emotion, making it possible to study human mental processes under realistic conditions such as visual noise and class imbalance.
Using the MOTOR-dataset, we established a zero-shot \textbf{benchmark} for state-of-the-art MLLMs and multi-agent systems. We also proposed \textbf{MOTOR-MAS}, a structured multi-agent framework that coordinates specialized agents through theory-driven sequential reasoning. On the MOTOR-Bench, MOTOR-MAS achieved a Macro-F1 score of \textbf{42.77}, outperforming the strongest single-model baseline by 15.93 points.

The absolute scores remain modest, which is expected. Naturalistic classroom data is noisier and more imbalanced than curated datasets, emotional expressions are subtle, and labels are skewed. The task is genuinely hard, and we see this difficulty as part of the benchmark's value.

This work still has several limitations. The current framework operates at the level of individual utterances and does not explicitly model the temporal dynamics of collaborative interaction. Second, our cue-based implementation relies on zero-shot large multimodal language models and therefore inherits their computational cost.  
Third, the MOTOR-dataset is collected from Finnish science education, representing one concrete instance of real-world collaborative interaction rather than a fully general setting. This choice was deliberate. High-quality expert annotation requires a well-established theoretical coding scheme, and the SRL framework provided exactly this foundation. The results demonstrate that grounding agent coordination in domain theory is an effective strategy for structured mental state inference. This principle is not specific to education. Other structured social contexts, such as clinical consultation or team-based problem-solving, have comparable theoretical frameworks. Applying this design strategy in those domains is a meaningful direction for future work.

\section*{Acknowledgments}

This work was supported by the Research Council of Finland project (grant 348765), flagship Profi 7 (grant 352788), EmotionAI (grants 336116, 345122, 359854), Research Fellow (371019), EU HORIZON-MSCA-SE ACMod (101130271), University of Oulu, and Infotech Oulu. We also wish to acknowledge CSC– IT Center for Science, Finland, for computational resources. Data collection was carried out with the support of the LeaF Research Infrastructure (https://www.oulu.fi/leaf-eng/), University of Oulu, Finland.

{
    \small
    \bibliographystyle{ieeenat_fullname}
    \bibliography{main}
}


\end{document}